\documentclass[lettersize,journal]{IEEEtran}

\usepackage{graphicx}
\usepackage{epstopdf}
\usepackage{amssymb}
\usepackage{amsmath} 
\usepackage{enumerate}
\usepackage{etaremune}
\usepackage{setspace}
\usepackage{comment}
\usepackage{color}
\usepackage[table]{xcolor}
\usepackage{soul,xcolor}
\setstcolor{red}
\usepackage{wrapfig}
\usepackage[algo2e]{algorithm2e} 
\usepackage{algorithm} 
\usepackage{algpseudocode} 
\usepackage[hidelinks]{hyperref}
\usepackage{cite}
\definecolor{blue0}{RGB}{100, 211, 100}
\definecolor{red}{RGB}{150, 11, 23}
\usepackage{soul}
\usepackage{booktabs}
\usepackage{float}
\usepackage{amsmath}
\usepackage{subfigure}
\usepackage{pgfplots} 
\pgfplotsset{compat=newest} 
\pgfplotsset{plot coordinates/math parser=false} 
\newlength\figureheight 
\newlength\figurewidth 
\usetikzlibrary{mindmap}
\usepackage{placeins}
\usetikzlibrary{arrows}
\usepgfplotslibrary{patchplots}
 \usepackage{soul}
 \usepackage{tcolorbox}
\newtcolorbox{mybox}{colback=yellow!50!white,colframe=yellow!100!black}
 \usepackage[switch]{lineno}

\def\BibTeX{{\rm B\kern-.05em{\sc i\kern-.025em b}\kern-.08em
    T\kern-.1667em\lower.7ex\hbox{E}\kern-.125emX}}

\begin{document}
\title{A Hybrid Task-Constrained Motion Planning for Collaborative Robots in Intelligent Remanufacturing}
\author{Wansong Liu$^{1}$, Chang Liu$^{2}$, Xiao Liang$^{3}$, Minghui Zheng$^{2}$
    \thanks{This work was supported by the USA National Science Foundation  (Grants: 2026533/2422826 and 2132923/2422640). This work involved human subjects or animals in its research. The authors confirm that all human/animal subject research procedures and protocols are exempt from the University at Buffalo's review board approval.}
	\thanks{$^{1}$ Wansong Liu is with the Mechanical and Aerospace Engineering Department, University at Buffalo, Buffalo, NY 14260, USA. {\tt\small Email: wansongl@buffalo.edu}.}%
	\thanks{$^{2}$ Chang Liu and Minghui Zheng are with J. Mike Walker '66 Department of Mechanical Engineering, Texas A\&M University, College Station, TX 77843, USA. {\tt\small Email: \{changliu.chris, mhzheng\}@tamu.edu}.}
    \thanks{$^{3}$ Xiao Liang is with Zachry Department of Civil and Environmental Engineering, Texas A\&M University, College Station, TX 77843, USA. {\tt\small Email: xliang@tamu.edu}.}%
	\thanks{$^*$ Correspondence to Minghui Zheng and Xiao Liang.}
}

\maketitle

\begin{abstract}
Industrial manipulators have extensively collaborated with human operators to execute tasks, e.g., disassembly of end-of-use products, in intelligent remanufacturing. A safety task execution requires real-time path planning for the manipulator's end-effector to autonomously avoid human operators. This is even more challenging when the end-effector needs to follow a planned path while avoiding the collision between the manipulator body and human operators, which is usually computationally expensive and limits real-time application. This paper proposes an efficient hybrid motion planning algorithm that consists of an $A^*$ algorithm and an online manipulator reconfiguration mechanism (OMRM) to tackle such challenges in task and configuration spaces respectively. The $A^*$ algorithm is first leveraged to plan the shortest collision-free path of the end-effector in task space. When the manipulator body is risky to the human operator, our OMRM then selects an alternative joint configuration with minimum reconfiguration effort from a database to assist the manipulator to follow the planned path and avoid the human operator simultaneously. The database of manipulator reconfiguration establishes the relationship between the task and configuration space offline using forward kinematics, and is able to provide multiple reconfiguration candidates for a desired end-effector's position. The proposed new hybrid algorithm plans safe manipulator motion during the whole task execution. Extensive numerical and experimental studies, as well as comparison studies between the proposed one and the state-of-the-art ones, have been conducted to validate the proposed motion planning algorithm. 
\end{abstract}


\section{Introduction}

\IEEEPARstart{I}{n} recent years, the rapid development of intelligent remanufacturing enables human-required repetitive and dangerous tasks to receive assistance from industrial manipulators. The human-manipulator collaborative tasks, e.g., disassembly of end-of-use products \cite{lee2024review,lee2022task,lee2022robot}, need them to work side-by-side in a sharing environment. When executing tasks, the proximity between human workers and manipulators brings potential collisions since manipulators usually move fast and human motion is usually dynamic with uncertainties \cite{liu2022dynamic,eltouny2024tgn}. Therefore, it's important to develop a reliable motion planning algorithm for manipulators to guarantee human workers' safety. 

Most robot-engaged task executions require planning a sequence of waypoints that enables the robot to safely move from an initial position to a goal position in the task space. To this end, extensive path planning algorithms have been proposed to generate a collision-free path for robots. The manipulator also needs to preserve task constraints throughout the planned motion in many real-world tasks, e.g., avoiding collision with human operators as well as following a desired task path \cite{zhao2020contact,liu2023task} or maintaining a desired end-effector's orientation \cite{stilman2007task}.
Such task-constrained motion planning problems usually are solved in the configuration space by finding the configuration that satisfies the constraints \cite{pan2015efficient}. The configuration space usually has higher dimension compared to the task space, especially for redundant manipulators, which limits the application of the mentioned grid methods and intelligent bionic planning methods due to the high computational cost.

To execute tasks as well as satisfy constraints successfully, many efforts have been devoted to planning the manipulator motion in the configuration space. Virtual potential field methods enable the manipulator to avoid obstacles and track references by creating repulsive and attractive forces on the manipulator respectively. For example, the joint velocity according to the repulsive and attractive forces is calculated in \cite{wang2018improved} to guide the manipulator to avoid the obstacle and follow the trajectory step by step. The safe set of the collision avoidance algorithm introduced in \cite{lin2017real} further reduces the counteraction between the repulsive-based and attractive-based velocity components.

Moreover, sampling-based methods keep choosing samples randomly or determinately in the configuration space and verifying the satisfaction of the desired constraints, and a continuous motion will be generated by connecting the admissible configurations. For example, the probabilistic roadmap method (PRM) \cite{kavraki1996probabilistic} generates random samples in the configuration space of the robot, and connects the generated free configurations by a fast local planner. A roadmap is constructed to handle multiple queries problems of motion planning. Rapid exploring random tree (RRT) and its modifications formed a sampling-based exploring tree in the configuration space and used constraint-satisfied samples to extend the branches until reach the target configuration \cite{cefalo2013task,wei2018method,oriolo2005motion,kingston2018sampling}. 

The aforementioned sampling-based planner have the capability of solving high degree-of-freedom (DOF) motion planning problems, but the generated trajectories tends to be not optimal. In recent years, some asymptotically optimal sampling-based algorithms, e.g., RRT* \cite{karaman2010incremental} and PRM* \cite{karaman2011sampling}, have been developed to generate not only feasible but also asymptotically optimal robot trajectories. Every newly generated configuration and its nearby configurations are employed to check if the planned path can be further shortened. Although the generated trajectories are globally optimal, the computational complexity of the asymptotically optimal sampling-based algorithms is high, and these algorithms may be not competitive for scenes with moving obstacles.

The manipulator motion planning can also be formulated as a nonlinear optimization problem that can be solved to obtain a series of robot motions while satisfying several constraints \cite{ratliff2009chomp,schulman2014motion}. As most of the formulated planning problems are highly non-convex and nonlinear, usually they are transformed to sequences of convex sub-problems, which can be solved iteratively \cite{reynoso2016convex,lin2018fast}. However, most of them are still subject to computational cost especially when there are moving obstacles changing the free configuration space in a complex way.

Recently, several motion planning studies aim to optimize the balance between two criteria, efficiency, and safety, especially in human-robot collaborative environments. For example, \cite{faroni2022safety} embedded a human model in the robot's motion planner and converted the robot's safety speed limit into configuration-space cost functions that drive the path’s optimization. \cite{flowers2023spatio} presents a spatio-temporal avoidance of predictions-prediction and planning framework (STAP-PPF) which proactively determines time-optimal robot paths by considering the predicted future human motions and robot speed restrictions. \cite{laha2023s} introduced an S* algorithm that leverages a graph search guided by an informed cost balance criterion. While these studies improve the planning efficiency in human-robot collaborative environments, the task constraints are not explicitly considered.

Actually, problems in the task space have to be converted into the configuration space online to find a locally optimal configuration as well as satisfying task-space-based constraints in the aforementioned optimization-based studies, e.g., \cite{reynoso2016convex,rubagotti2019semi,nubert2020safe}. Such a conversion is inevitable and may increase the computational cost, especially when facing moving obstacles. The moving obstacle makes the manipulator motion is more difficult to be efficiently planned since the free configuration space is changing with the obstacle state in real-time. In this paper, we present a hybrid algorithm to plan safety manipulator motion in a dynamic environment. A collision-free path is first planned as the end-effector's reference in the task space. The manipulator body's motion safety is then ensured by OMRM in the configuration space when facing a moving obstacle as well as following the planned path. OMRM uses the alternative manipulator configuration provided by a database to satisfy nonlinear constraints in motion planning. Instead of converting the task space to the configuration space online, the database of the manipulator reconfiguration establishes the relationship between the task space and configuration space offline, which significantly reduces the computational cost. 

\begin{figure}[ht]
	\centering 
	\includegraphics[scale=0.4]{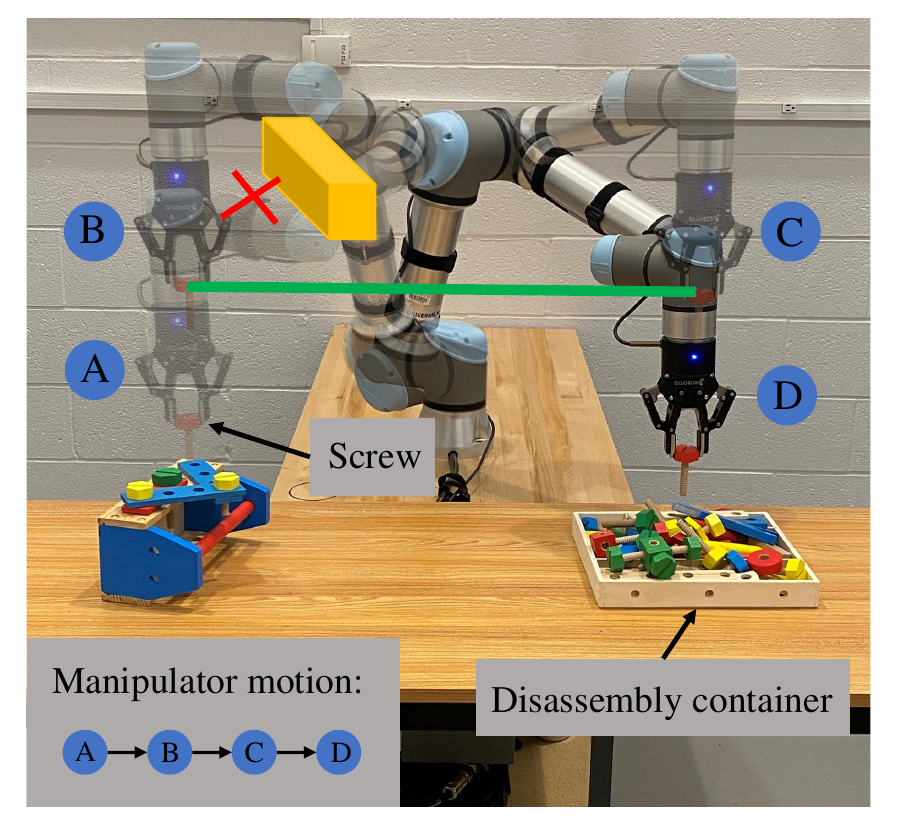}
	\caption{The potential collision scene: the manipulator motion is introduced in the left-down corner of the figure, the orientations of the end-effector in the four manipulator states are the same, and we focus on the motion from state B to state C. The green line is the planned collision-free path, the yellow box is the obstacle.} \label{fig:use-case}
\end{figure}

The main contributions of this work are summarized as follows. (1) We presented a new hybrid motion planning algorithm which aims to efficiently plan the shortest end-effector's path while avoiding collisions between the whole manipulator and the dynamic environment with minimum reconfiguration effort (i.e., minimum joint-angle change). (2) The database of the manipulator's reconfiguration minimizes the necessity of the online task-configuration-space conversion, which improves the motion planning efficiency. (3) The effectiveness of the proposed motion planning algorithm has been experimentally validated. 

The remainder of this paper is organized as follows. Section 2 gives the overview of the proposed motion planning algorithm; Section 3 describes the end-effector path planning in the task space; Section 4 presents the avoidance of collision between the manipulator body and the human operators in the configuration space; Section 5 shows the effectiveness of the proposed algorithm in the simulation; Section 6 demonstrates the experimental tests with two scenarios; Section 7 concludes this paper. 

\section{Overview of the proposed motion planning algorithm}

This section presents the overview of the proposed new hybrid algorithm for manipulator motion planning. This algorithm aims to efficiently generate (1) a collision-free path for the end-effector with the shortest distance from the start position to the goal position in the task space, and (2) a collision-free path for the manipulator links with minimum reconfiguration effort. We first briefly introduce a potential collision scene between the obstacle and the manipulator body.
We then formulate the planning problem with collision avoidance into an optimization problem with constraints in both task and configuration spaces. Eventually, we focus on how the optimization problem is solved using the proposed motion planning algorithm that consists of hybrid online-offline reconfiguration.

Fig.~\ref{fig:use-case} illustrates a potential collision scene. Suppose the manipulator is conveying components, e.g., screws, to a container in a disassembly process. The end-effector of the manipulator carrying a screw tries to move from state B to state C with the shortest distance. The task completion requires the end-effector to follow the green obstacle-free path. Meanwhile, an obstacle represented by the yellow box appears, which would cause a collision if the manipulator moves forward following the planned path. In this case, to follow the planned path and avoid such a collision simultaneously, the configuration causing a collision has to be substituted by a new collision-free configuration leading to the same end-effector's position.

We first introduce needed notations and definitions. The manipulator end-effector's position is denoted as $x$ such that $x \in \mathcal{X} \subset \mathbb{R}^{3}$, where $\mathcal{X}$ is the task state in $3$-dimensional task space. The manipulator configuration is denoted as $\theta$ such that $\theta \in \Theta \subset \mathbb{R}^{q}$, where $\Theta$ is the configuration state in $q$-dimensional configuration space. The area occupied by the manipulator with the configuration $\theta$ in the task space is represented as $\mathcal{M}(\theta) \subset \mathbb{R}^{3}$. The calculation of the occupied area is based on the forward kinematics of the manipulator. It considers the length and radius of individual manipulator links, as well as the precise positional offset associated with manipulator joints. The relationship between the end-effector position $x$ and the manipulator configuration $\theta$ is based on the forward kinematics of the manipulator, e.g.,
\begin{equation}
    x=F(\theta)
    \label{FK_first}
\end{equation}
where $F \in \digamma$ denotes the forward kinematics.
In order to meet the safety requirement, the human operator, which can be treated as a moving obstacle, has to be considered in the manipulator motion planning.
The area occupied by the human operators in the task space is $\mathcal{O}^T \subset \mathbb{R}^{3}$. What's more, considering the whole manipulator body, the dangerous manipulator state is marked as $\mathcal{O}^J \subset \mathbb{R}^{q}$, in which $\mathcal{O}^J$ is defined as:
\begin{equation*}
    \forall \theta \in \mathcal{O}^J,~\mathcal{M}(\theta) \cap \mathcal{O}^T \neq \emptyset
\end{equation*}
We define the safe task state and the manipulator configuration state respectively as follows:
\begin{itemize}
    \item The obstacle free end-effector state in task space:
    \begin{equation}
        \mathcal{X}^{free}=\mathcal{X} \setminus \mathcal{O}^T
    \end{equation}
    \item The obstacle free manipulator configuration state in configuration space:
    \begin{equation}
        \Theta^{free}=\Theta \setminus \mathcal{O}^J
    \end{equation}
\end{itemize}

As shown in Fig.~\ref{fig:use-case}, assuming the robot needs to plan $K$ steps in total to move from the initial state B to the target state C, the end-effector's position at step $k$ is marked as $x_k \in \mathcal{X}$, and the end-effector path from the initial position $x_0$ to the goal position $x_{K+1}$ is represented as
$[x_0,x_1,x_2\dots,x_K,x_{K+1}] \in \mathcal{X}^{K+2}$. 
We do not consider the orientation of the manipulator in this study, and
the manipulator motion planning aiming to generate a shortest collision-free path can be formulated into the following optimization problem:

\begin{subequations}\label{constraint}
\begin{align}
\min_{\hat{x}, \hat{\theta}}& \quad  \sum_{k=1}^{K+1}{\left\|x_{k}-x_{k-1}\right\|}\\ \label{1st_constraint}
\textrm{s.t.} & \quad x_{k} \in \mathcal{X}^{free}_{k} \subset \mathbb{R}^{3}\\ \label{2nd_constraint}
  &\quad \delta_{L} \leq \left\|x_{k}-x_{k-1}\right\| \leq \delta_{U}    \\ \label{3rd_constraint}
  &\quad x_{k}=F(\theta_{k})\\ \label{4th_constraint}
  &\quad \theta_{k} \in \Theta^{free}_{k} \subset \mathbb{R}^{q}
\end{align}
\end{subequations}
where $\delta_{L}$ and $\delta_{U}$ are the minimum and maximum distances between two successive waypoints, $\theta_k$ is the manipulator joint configuration at step $k$, and $\hat{x}=[x_1,x_2\dots,x_K]$ and $\hat{\theta}=[\theta_1,\theta_2\dots,\theta_K]$ are planned manipulator states in task space and configuration space, respectively. Note that $\mathcal{X}^{free}_k$ and $\Theta^{free}_k$ may change at each step based on the moving human operator's state.

Eq.~(\ref{1st_constraint}) indicates that the planner needs to guarantee the end-effector's safety in the task space. To enable the task to be executed efficiently and safely, 
Eq.~(\ref{2nd_constraint}) is included to limit the distance between waypoints to a certain range. To guarantee that the joint configurations match up the end-effector's position, the nonlinear mapping Eq.~(\ref{3rd_constraint}) is included. To make sure that the whole manipulator body is collision-free, Eq.~(\ref{4th_constraint}) is included. Note that although Eq.~(\ref{4th_constraint}) implies Eq.~(\ref{1st_constraint}), Eq.~(\ref{1st_constraint}) and Eq.~(\ref{4th_constraint}) are guaranteed by $A^*$ in the task space and OMRM in the configuration space, respectively. Therefore, we include both of them in the formulation.

\begin{figure*}[t]
	\centering 
	\includegraphics[scale=0.49]{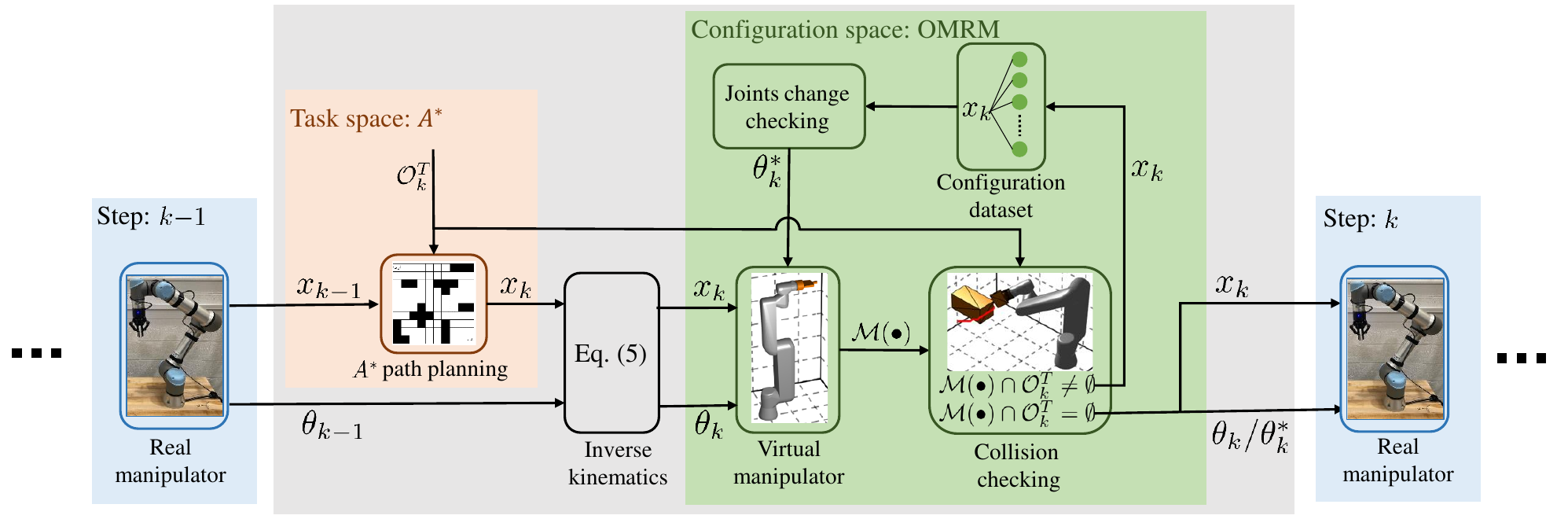}
	\vspace{3pt}
	\caption{The framework of solving the formulated optimization problem, where the green dots represent the joint configuration candidates corresponding to $x_k$ 
	} \label{fig:formulation}
\end{figure*}

The manipulator motion generation is regarded as solving the above optimization problem. Actually, the drawback of the traditional task-constrained manipulator motion generation lies in that the online conversion between the task and configuration spaces is inevitable. It's difficult to solve an optimization problem efficiently, especially when considering the performance of the task execution and the collision avoidance of the whole manipulator body simultaneously. Therefore, in this paper, we propose a new hybrid online-offline manipulator motion planning algorithm to solve the optimization problem Eq.~(\ref{constraint}) and respectively satisfy the desired constraints in the task space and configuration space in the following ways:
\begin{itemize}
    \item The manipulator workspace is converted to a 3-dimensional (3D) grid map in real-time. In order to satisfy Eq.~(\ref{1st_constraint}) in the task space, $A^*$ algorithm is applied to eliminate the nodes which contain the human operator and plan a path of the end-effector only in $\mathcal{X}^{free}$.
    \item The waypoints of the planned end-effector path is generated between two successive nodes. The node size limits $\left\|x_{k}-x_{k-1}\right\|$ such that Eq.~(\ref{2nd_constraint}) is satisfied.   
    \item  Eq.~(\ref{3rd_constraint}) describes the highly nonlinear mapping from the joint configuration to the end-effector's position. The manipulator motion usually is controlled in the configuration space. 
    In order to find the configuration $\theta_k$ such that the manipulator can reach the desired waypoint $x_k$, the traditional way is using the inverse kinematics algorithm: 
    \begin{equation}
           \theta_{k}=J^{\dagger}(\theta_{k-1})(x_{k}-F(\theta_{k-1}))+\theta_{k-1}
    \end{equation}
 
    where $J^{\dagger}(\theta_{k-1})\in \mathbb{R}^{q \times 3}$ is the pseudo inverse of the manipulator Jacobian evaluated at $\theta_{k-1}$. In the case of $q {>} 3$, inverse kinematics usually provides one optimization-based configuration solution for a desired $x_k$ since the manipulator is kinematically redundant. However, obtaining $\theta_k$ online is difficult if the collision checking is integrated into the inverse kinematics. We employ forward kinematics to construct a joint configuration database offline. Such a database provides multiple joint configurations which lead to the desired $x_k$. Therefore, instead of directly solving for one collision-checked configuration $\theta_k$ online, multiple configuration candidates for the desired $x_k$ are pre-computed offline, and Eq.~(\ref{3rd_constraint}) is satisfied instinctively.     
    \item Eq.~(\ref{4th_constraint}) requires that the manipulator body cannot have any collisions with the human operators. We develop an online manipulator reconfiguration mechanism (OMRM) to satisfy this constraint in the configuration space. If $\mathcal{M}(\theta_k) \cap \mathcal{O}^T_k \neq \emptyset$ in the current step $k$, OMRM would select one optimal joint configuration $\theta^*_k$ from the database based on the desired $x_k$, the reconfiguration effort, and the collision checking.  
    The optimal joint configuration $\theta^*_k$ replaces $\theta_k$ such that $\mathcal{M}(\theta^*_k) \cap \mathcal{O}^T_k = \emptyset$.
\end{itemize}

By solving the formulated optimization problem in real-time, the end-effector follows the shortest planned path and the manipulator body has no collision with the human operators during the whole task execution. Fig.~\ref{fig:formulation} illustrates that how the optimization is solved step by step. The $A^*$ path planning algorithm handles the generation of the end-effector's each waypoint $x_k$ in the task space. If the joint configuration $\theta_k$ obtained from a real manipulator faces a collision with the human operators, OMRM would handle the manipulator reconfiguration in the configuration space to avoid the human.

\section{End-effector path planning in task space}

This section presents details on how the end-effector path is planned. We consider planning a shortest end-effector path from the initial position $x_0$ to the goal position $x_{K+1}$ in the task space. Here we leverage $A^*$ algorithm  \cite{hart1968formal} to obtain such a path in a dynamic environment.

Firstly, the workspace of the manipulator is rasterized, and 3D grid cells are generated. The node is at the center of the cell and stands for the end-effector's position $x$ in the task space.
Then, if the end-effector is in the current node $x_{k-1}$, the next waypoint of the end-effector, i.e., $x_k$, will be searched from the feasible successor nodes which have no overlap with any humans. For each reachable and feasible node $n$, the next node $x_k$ is selected based on the shortest distance which is defined as following equation:

\begin{equation} \label{distancecost_new}
f(n)=g(n)+h(n)
\end{equation}
where $f(n)$ stands for the total distance cost from $x_0$ to $x_{K+1}$ passing a node $n$, $h(n)$ is the euclidean distance cost from the node $n$ to $x_{K+1}$, and $g(n)$ denotes the actual distance cost from $x_0$ to the node $n$ through the planned path $[x_1,\dots,x_{k-1}]$ with the following equation:
\begin{equation}
    g(n)=g(x_{k-1})+\Delta g(x_{k-1},n)
\end{equation}
where $g(x_{k-1})$ is the actual distance cost of the planned path, and $\Delta g(x_{k-1},n)$ is the distance cost from the node $x_{k-1}$ to node $n$. The node with the minimum total distance cost is chosen as $x_k$ until the end-effector reaches the goal node $x_{K+1}$.

In a dynamic environment, the grid map is updated with a certain frequency in real-time such that the area occupied by the human operators $\mathcal{O}^T_k$ is informed to the end-effector path planning.
$x_{start}$ is the local start node and is updated iteratively. $Open$ list contains all obstacle-free successor nodes. $\hat{f}$ list contains the cost function values of obstacle-free successor nodes.
The moving obstacle first is treated to be relatively static to plan a local path $Closed$ from $x_{start}$ to $x_{K+1}$. Every waypoint of the local path is determined based on the cost function value in $\hat{f}$. 
Next, based on a certain step size $\gamma$ determined by the map updating frequency, we re-plan the local path and construct the global planned path $\hat{x}=[x_1,x_2\dots,x_K]$ for the end-effector from $x_0$ to $x_{K+1}$.

\section{Manipulator reconfiguration in configuration space}
A collision-free path is planned for the end-effector to reach the goal position safely. Actually, safety is guaranteed only for the end-effector in the task space, and the collision between the human and the manipulator body still may happen due to an improper joint configuration.
This section presents the avoidance of collision between the whole manipulator body and the human operators. 
We construct a database based on Eq.~(\ref{FK_first}) to provide multiple manipulator configurations for a desired $x_k$, and develop OMRM to achieve the collision avoidance for the whole manipulator body.

\subsection{Construction of manipulator configuration database}
\begin{figure}[t]
	\centering 
	\includegraphics[scale=0.28]{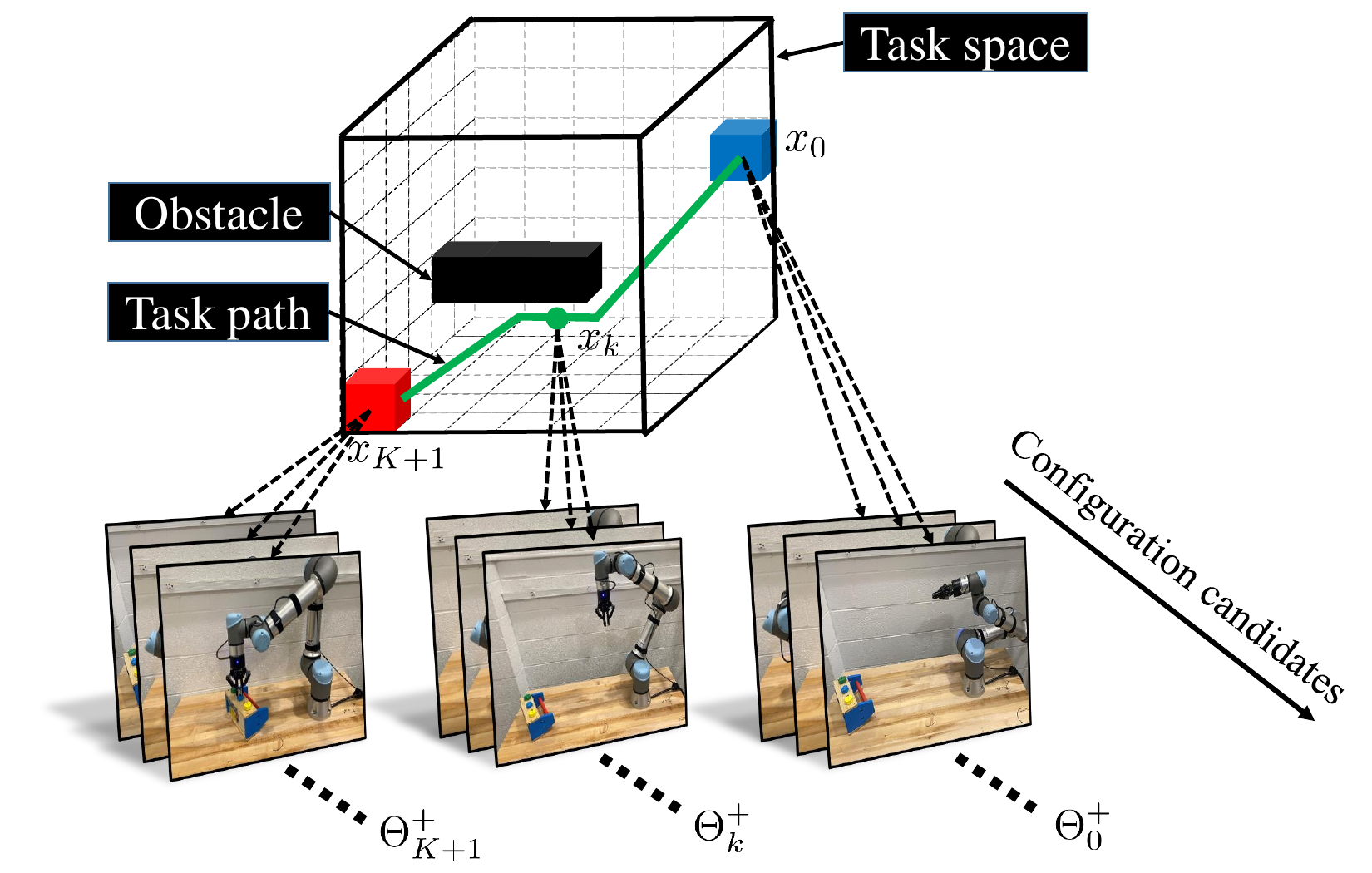}
	\vspace{5pt}
	\caption{The structure of the reconfiguration database: each $\Theta^+$ stands for the set of configuration candidates that leads to the same $x$ in the task space.} \label{fig:database}
\end{figure}

This subsection describes the details of the database construction. In most manipulator motion planning algorithms, a desired joint configuration, i.e., $\theta_k$ for a specific end-effector's position, i.e., $x_k$, is solved by the optimization-based inverse kinematics \cite{rakita2017motion,sinha2019geometric,rakita2018relaxedik}. On the other hand, the computation time of solving the inverse kinematics may increase if the collision checking of the whole manipulator body is embraced, especially for redundant manipulators. Moreover, the inverse kinematics may fail to converge a valid solution and usually provides only one configuration solution. Actually, a redundant manipulator has the property that it has the different (infinite) joint configurations to reach one single end-effector's position. Therefore, we take advantage of such a property using forward kinematics to construct a database of the manipulator configuration offline. The database provides a set of manipulator configurations $\Theta^+_k$  for a single $x_k$, where $\Theta^+_k = [\theta_1, \theta_2, \dots, \theta_c, \dots, \theta_C] \in \mathbb{R}^{q{\times}C}$, $c$ is the index, and $C$ is the total number of the configuration candidates and may change based on different $x_k$. Fig.~\ref{fig:database} presents the structure of the database. By selecting a proper configuration as $\theta_k$ from $\Theta^+_k$, the expensive cost of computing a desired configuration is transformed to a cheap selection cost.

The database of the manipulator reconfiguration is generated based on forward kinematics. Kinematic equations is used to compute the end-effector's position using the manipulator parameters, e.g., the joint angle and the manipulator link's length. Suppose a manipulator is characterized by a sequence of $q$ links $S_i$, $i=1,2,\dots,q$, one for each joint angle $\vartheta_i$ in the robot. Each joint has a local coordinate system, e.g., $x_i$, $y_i$, and $z_i$. The manipulator link $S_i$ is regarded as the coordinate transformation between tow successive joints. The rotational and translational transformations are represented as the Denavit-Hartenberg (DH) matrix:
\begin{equation}\label{EQ:DHtrans}
\setlength{\arraycolsep}{1.2pt}
    \textbf A_i = \left[\begin{array}{cccc}\cos{\vartheta_{i}} & -\sin{\vartheta_{i}} \cos{\alpha_{i}} & \sin{\vartheta_{i}} \sin{\alpha_{i}} & a_{i} \cos{\vartheta_{i}} \\ \sin{\vartheta_{i}} & \cos{\vartheta_{i}} \cos{\alpha_{i}} & -\cos{\vartheta_{i}} \sin{\alpha_{i}} & a_{i} \sin{\vartheta_{i}} \\ 0 & \sin{\alpha_{i}} & \cos{\alpha_{i}} & d_{i} \\ 0 & 0 & 0 & 1\end{array}\right]
\end{equation}
where the $\vartheta_i$ is the manipulator joint describing as the angle change from $x_{i-1}$ to $x_i$ about $z_{i-1}$, the $\alpha_i$ is angle change from $z_{i-1}$ to $z_i$ about $x_i$ , $d_i$ is the offset between two joints along $z_{i-1}$, and $a_i$ is the offset between two joints along $x_i$. The transformation of the last link $\textbf T_{e}$ is obtained with the following equation: 
\begin{equation}\label{EQ:ForwardK_new}
    \textbf T_{e} = \prod_{i=1}^{q}{\textbf A_i}
\end{equation}
where the translation part of $\textbf T_{e}$ is the end-effector's position. The $a_i$, $d_i$, and $\alpha_i$ values depend on the manipulator type. 
Abundant configurations are first generated by specifying the joint limit and changing the joint angle $\vartheta_i$ with a certain interval $\eta$.
The $\eta$ value determines the total number of generated configurations.
Next, the corresponding end-effector's positions are computed using Eq.~(\ref{EQ:ForwardK_new}). 
Finally, considering the database is required to provide the configuration candidate set $\Theta^+$ based on a certain $x$, the configurations that lead to the end-effector's position $x$ within a task space error tolerance $\zeta$ are classified as the elements of $\Theta^+$, i.e.,
\begin{equation*}
   \forall \theta \in \Theta^+,~|| F(\theta)-x||_2 \leq \zeta  
\end{equation*}
Note that the error tolerance $\zeta$ is a user-defined value and is closely aligned with the grid size of the waypoints stored within the database. It affects the number of configuration candidates for each waypoint.

\subsection{OMRM}
This subsection describes how OMRM selects the optimal configuration $\theta_k^*$ from the database. When facing a collision of the manipulator body, the database has the capability to provide a set of configuration candidates $\Theta^+_k$ for a desired $x_k$ such that any configuration candidate $\theta_c$ from $\theta_k^+$ can lead to $x_k$. On the other hand, not every configuration candidate can make the manipulator avoid the human. A straightforward way to guarantee the human to be avoided would be first checking collision for all configuration candidates and then randomly selecting a collision-free configuration to replace the current one. However, this approach is inefficient and the selected configuration may cause large joint angle changes. Therefore, we develop OMRM to efficiently select a configuration $\theta_k^*$ with the minimum joint angle change from $\Theta^+_k$ to make the manipulator body to avoid the human.

The configuration selection procedure of OMRM is
based on two rules which are defined as follows:
\begin{itemize}
    \item \textit{Rule 1:} The joint angle change is preferred to be minimized during the avoidance.
    \item \textit{Rule 2:} The manipulator with the selected configuration can not have any collision with the human operators.
\end{itemize}

When a manipulator with configuration $\theta_k$ reaches the position $x_k$ and collides with the human $\mathcal{O}^T_k$, OMRM finds the candidate configuration set $\Theta^+_k$ from the database first, delaying any collision checking, and then ranks the elements of $\Theta^+_k$ based on the root mean square error (RMSE) compared to $\theta_k$, finally, the elements of the ranked $\Theta^+_k$ is checked collision successively. A collision between a manipulator and a human is defined as:
 \begin{equation}
    d(\mathcal{M}(\theta),\mathcal{O}^T) \leq 0
\end{equation}
where the manipulator and the human are respectively represented using the corresponding mesh model and cylinder model. Additionally, $d(\bullet)$ leverages the flexible collision library \cite{pan2012fcl} to compute the minimum distance between two models and determine whether two models have any overlaps in the task space. Note that one potential limitation of our proposed hybrid planning approach is that minimizing joint-angle changes may bring the manipulator into close proximity with obstacles, including humans. To guarantee a minimum distance between the manipulator and any obstacle, we incorporated a safety distance into our collision checking process. Specifically, when the minimum distance between the robot and a human is smaller than a predefined safety distance, we treat it as a collision. This strategy effectively minimizes reconfiguration efforts while maintaining a minimum distance between the human and the manipulator.

\textbf{Algorithm~\ref{selection_new}} describes the details of the proposed OMRM. Note that the following case may happen: $A^*$ finds a path for the end-effector within a narrow environment, but the physical body of the robot may encounter a collision. When none of the configuration candidates stored in the database is suitable for the obstacle avoidance, instead of re-running A*, we claim that the planning directly fails and, instead, we use traditional sampling-based method for the next-step planning.

In theory, the number of the configuration candidates and the corresponding manipulator states are finite for a specific $x_k$, thus OMRM is guaranteed to provide an alternative configuration if one exists. The guarantee does not extend to the overall planner, as $A^*$ may successfully find a collision-free path but OMRM may fail to provide collision-free solutions. Furthermore, OMRM does indeed possess resolution completeness. The discretization of potential configuration candidates significant influences the solutions when facing a potential collision.

\begin{algorithm}[h]
    $\bullet$ Obtain $\mathcal{O}^T_k$ based on the human's position. \\
	$\bullet$ Get $\Theta^+_k$ from the database based on $x_k$.\\
    $\bullet$ Calculate RMSE of each element in $\Theta^+_k$. \\
    $\bullet$ Rank $\Theta^+_k$ based on the RMSE from the minimum to the maximum. \\
	$\bullet$ Set iteration number $j = 1$. \\
	$\bullet$ Assign the $j$ th element of the ranked $\Theta^+_k$ as a temporal $\theta^*_k$.\\
    \While{$d(\mathcal{M}(\theta^*_k),\mathcal{O}^T_k) \leq 0$}
    {
    $\bullet$ Increment $j$. \\
    \eIf{$j \leq C$}
                {$\bullet$ Assign the $j$ th element of the ranked $\Theta^+_k$ as a temporal $\theta^*_k$.}
                {$\bullet$ No alternative configuration exist.}
    }
    $\bullet$ Return the optimal manipulator configuration $\theta_{k}^{*}$.
    \caption{OMRM}
	\label{selection_new}
\end{algorithm}

\begin{figure*}[t]
	\centering 
	\includegraphics[scale=0.68]{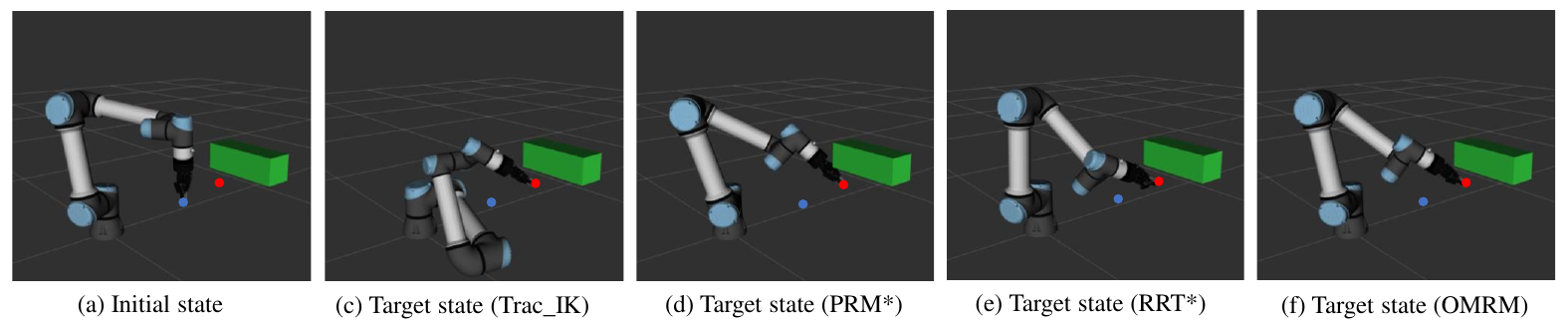}
	\caption{Planning simulation results: the blue dot is the start position, the red dot is the target position, the green box represents the obstacle.} \label{fig:sim_com}
\end{figure*}

\section{Simulation}

This section presents numerical studies to validate the effectiveness of OMRM. The simulation section consists of two parts, which are the simulation setup and the effectiveness evaluation of OMRM comparing with other planning methods.
\subsection{Simulation setup}
\subsubsection{Simulation platform}
We employ an open-source motion planning platform MoveIt based on the Robot Operating System (ROS) to validate the effectiveness of OMRM. 
All simulations were run on an Intel i9-12900K Processor. MoveIt construct a simulation model of the UR5e robot arm based on the corresponding Unified Robotics Description Format (URDF) file. The URDF file contains key kinematic information of the manipulator, which can be used to calculate the occupied area in the task space.
A simulation model of the UR5e robot arm with 6-DOF is applied in MoveIt. To guarantee the simulation effectiveness, the UR5e simulation model shares the same DH parameters with a real UR5e robot. 
To fully simulate the manipulator working state, a Robotiq 2F-85 gripper model with 17mm length has been attached to the UR5e model as the end-effector.
\subsubsection{Manipulator configuration database}
In the construction of the database used in OMRM, the joint range of the manipulator is specified from $-180^o$ to $180^o$. We split the joint range based on the interval $\eta=4^o$. It guarantees a sufficiently ample pool of configuration candidates for each designated waypoint, while simultaneously ensuring the efficient utilization of the database. The tolerance of the task space error $\zeta$ is manually selected to be $0.01m$.
This value is configured to guarantee comprehensive coverage of all necessary end-effector positions within the workspace.
To reduce the database size and the robot configuration selection time, the database only provides the first 5 UR5e joint angles since the $6$th joint angle doesn't affect the gripper's position. The original database has $90^5$ configurations in total. The configurations are first classified according to the corresponding end-effector's position and stored as cells. Considering the cost of the data storage and the lookup time, we specify the range of the end-effector's position to eliminate the unnecessary cells. The range of the end-effector's position in the task space is specified from 0.0m to 0.7m for all $X$, $Y$, and $Z$ directions. Eventually, the configurations of the final dataset can reach 357911 waypoints, and the average configuration candidates for each cell are more than 1000. The grid size for the waypoint in the database is selected to be 0.01m. This grid dimension aligns with the designed error tolerance $\zeta$. The database size, occupying 6.9GB of computer memory, is preloaded into the computational memory of the computer in preparation for both simulations and experimental trials. Once a re-configuration is requested, our OMRM is designed to directly select configuration candidates from the previously loaded computer memory. This strategy effectively eliminates the necessity of re-loading the database.

\subsection{Simulation results: comparisons with other methods}

This subsection presents the motion planning results using different methods. The selection of baseline methods stems from the following two primary rationales. (1) The methods used for the comparative analysis have been widely used in the realm of robot planning, particularly concerning robots with high degree of freedom. For example, both the Trac-IK and our OMRM can provide the robot configuration based on the end-effector's position, rendering them suitable for assessing planning efficiency through the comparative analysis. Moreover, the RRT* and PRM* planners can provide near-optimal robot configuration, making them apt choices for evaluating the trajectory optimility of the planning. (2) We are not particularly focusing on the safe and efficiency collaboration between human and robot; instead, we particularly focus on efficiently planning the ``shortest'' end-effector's path while avoiding collisions between the whole manipulator and the dynamic environment with ``minimum'' reconfiguration effort (i.e., minimum joint-angle change). In our case, human operators are considered as moving obstacles and no human models or human motion prediction will be Incorporated.

\subsubsection{Implementation details}

$x_0=(0.42m, 0.10m, 0.26m)$ is set to be the planning start position. The length, width, and height of the obstacle are 0.3m, 0.1m, and 0.1m, respectively. We simulate 5 different planning scenarios by changing the positions of the obstacle and the goal. In addition, the planning goal locates always below the obstacle.

Considering the asymptotically optimal sampling-based planners, i.e., PRM* and RRT*, tend to find the optimal solution as the planning time increases, we set a fixed planning time and quantify the re-configuration effort from the start position to the target position. The re-configuration effort is quantified upon the average joint change. To enable the asymptotically optimal sampling-based planners to find a feasible trajectory successfully and have extra time to optimize the planned trajectory, the planning time is defined to be 500ms. For each scenario, we run 10 tests for each method, and totally 200 solutions have been collected. 

\subsubsection{Motion planning results}

\begin{table}[!]
	\centering
	\begin{tabular}{cccc}
\toprule[1.5pt]		~  & Solution  & Joint Change & Env. \\
~ & Time (ms) & Effort (rads) &  Collisions
\\\toprule[1.5pt]
		 OMRM & 7.415 $\pm$ 1.689 & 0.533 $\pm$ 0.165  & 0 \\
		
		Trac-IK  & 0.107 $\pm$ 0.072 &  1.435 $\pm$ 0.436 & 28 \\
		
		PRM*  & 500.000 (46.522) & 0.728 $\pm$ 0.475 & 0\\ 
		
		RRT*  & 500.000 (36.150) & 0.817 $\pm$ 0.544 & 0\\
		\toprule[1.5pt]
	\end{tabular}
	\vspace{5pt}
	\caption{Summary of results from simulations}
	\label{tab:comparison_detail}
\end{table}

\begin{table}[!]
	\centering
	\begin{tabular}{cc}
\toprule[1.5pt]		Task  & Solution time/ms  \\\toprule[1.5pt]
		 Candidate set lookup upon waypoint & 1.820\\
		
		RMSE calculation  & 1.133\\
		
		Ranking candidates  & 0.178 \\
		
		Solution search upon collision-checking  & 4.284 \\
		\toprule[1.5pt]
		sum & 7.415 \\
		\toprule[1.5pt]
	\end{tabular}
	\vspace{5pt}
	\caption{Detail of OMRM computational process}
	\label{tab:OMRM_detail}
\end{table}

We evaluate the test results based on three primary measures: solution time (ms), joint change effort (rads), and the total number of environment collisions. Table~\ref{tab:comparison_detail} shows the summary of the simulation results. Particularly, besides presenting the fixed planning time of PRM* and RRT*, we also provide the average minimum time of finding a feasible solution for the two planners. Fig.~\ref{fig:sim_com} presents the initial state and target states of the robot using different kinds of planning methods. The comparison indicates a few points. First, compared to Trac-IK, the benefits of our OMRM, e.g., less joint change effort and collision avoidance, outweigh the cost of the extra computational time. Additionally, the Track-IK approach lacks the capability to factor in potential collisions involving either the environment or the manipulator itself.
The other point is that although we set extra time for PRM* and RRT* to optimize the solution, our OMRM still outperforms them regarding the joint change effort. And the prolonged and inefficient computational time associated with PRM* and RRT* significantly diminishes their practicality, particularly in scenarios involving dynamic obstacles.

\begin{figure*}[t]
	\centering 
 \fcolorbox{white}{white}{
	\includegraphics[scale=0.63]{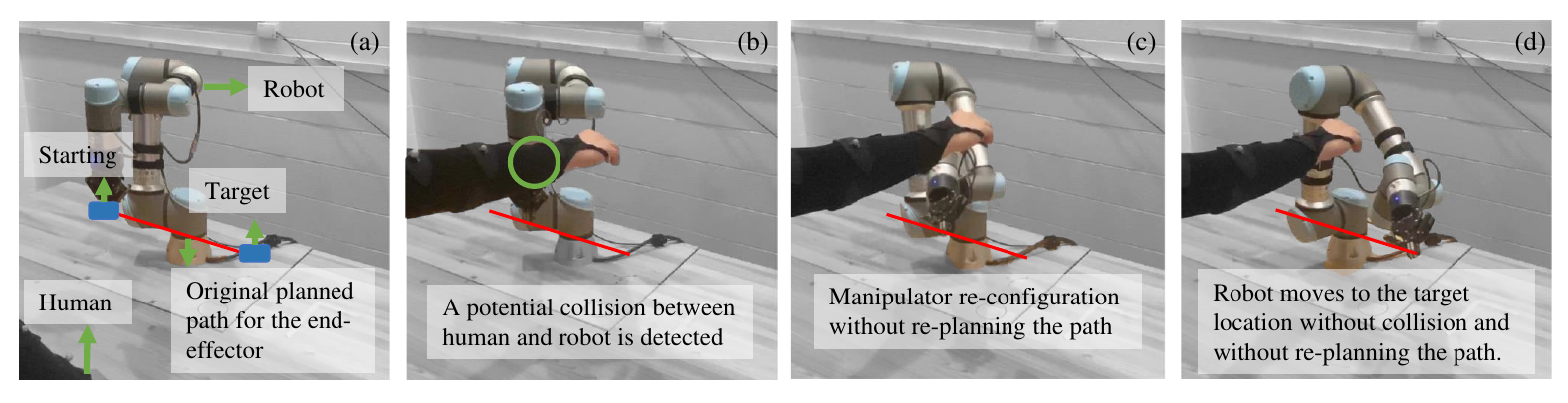}}
	\caption{UR5e motion planning experimental test scenario A: Sub-figure (a) is the initial frame of the task execution by the manipulator, and the red line is the end-effector's path planned by the A* algorithm. A sudden intervention occurs with the appearance of the human arm, blocking the manipulator's intended movement, as shown in (b). The green circle indicates the potential collision on the manipulator link. The manipulator changes its configuration and moves below the human arm to avoid the collision, as shown in (c). A short video of the experimental tests is available via this \href{https://zh.engr.tamu.edu/wp-content/uploads/sites/310/2024/06/OMRM_supplemental_videos.mp4}{{\underline{link}}} and in the supplemental materials.}  \label{fig:expA}
\end{figure*}

\begin{figure*}[t]
	\centering 
	\includegraphics[scale=0.485]{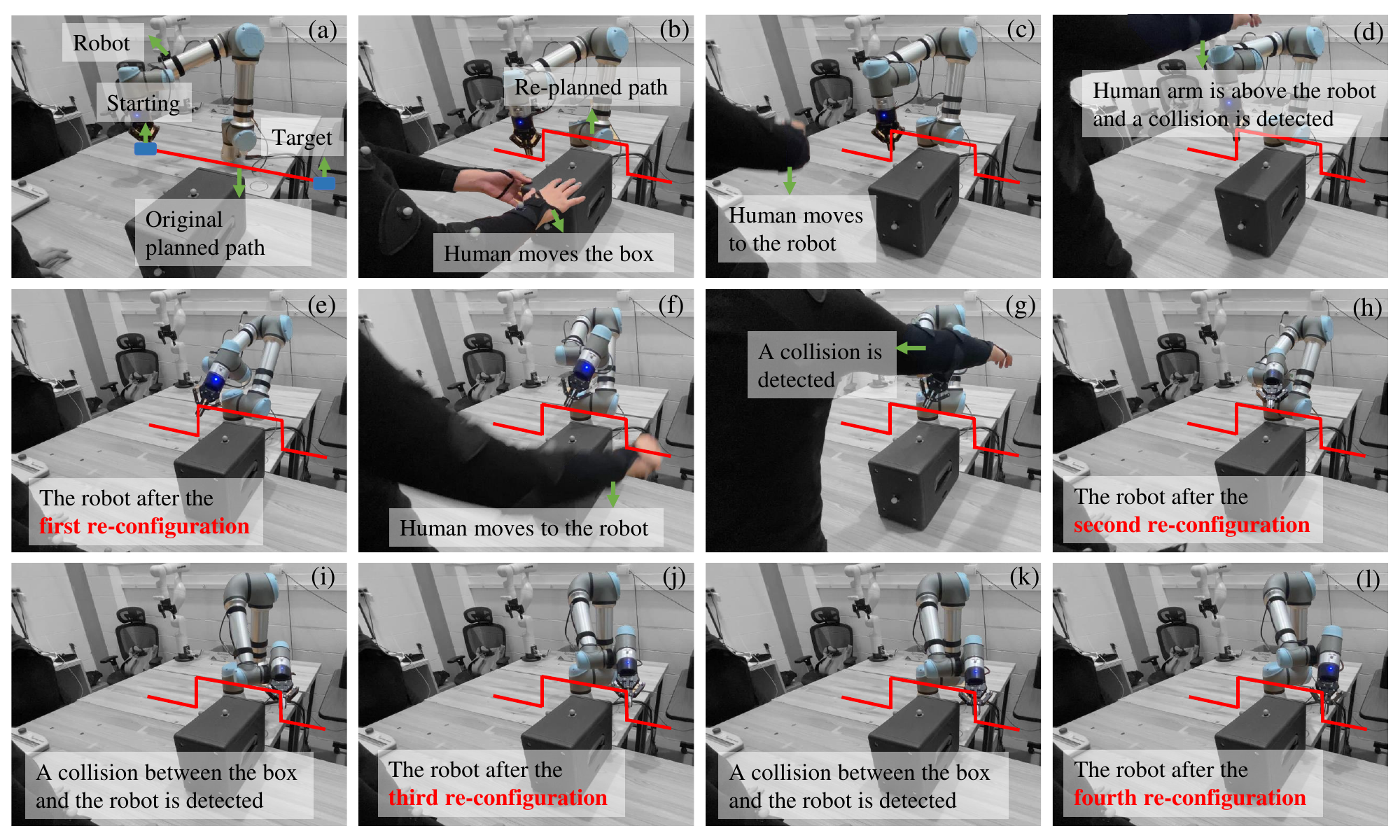}
	\caption{UR5e motion planning experimental test scenario B: Multiple intervention occurs with the movements of the human operator. Sub-figure (a) is the initial frame of the task execution by the manipulator, where the red line is the end-effector's path planned by the A* algorithm. The human operator intervenes by relocating a box, obstructing the original planned path. As such, the A* algorithm re-plans the path based on the box's position, as shown in (b). The human blocks the manipulator's intended movements, which triggers the first and second re-configurations, as shown in (d) and (g). In addition, (i) and (k) illustrate the moments when collisions between the box and the manipulator are detected. These collisions activate the third and fourth re-configurations. A short video of the experimental tests is available via this \href{https://zh.engr.tamu.edu/wp-content/uploads/sites/310/2024/06/OMRM_supplemental_videos.mp4}{{\underline{link}}} and in the supplemental materials.} \label{fig:expB}
\end{figure*}

\begin{figure*}[t]
	\centering 	
	\subfigure[The position of the end-effector in the X direction]{
	\includegraphics[width=0.62\columnwidth]{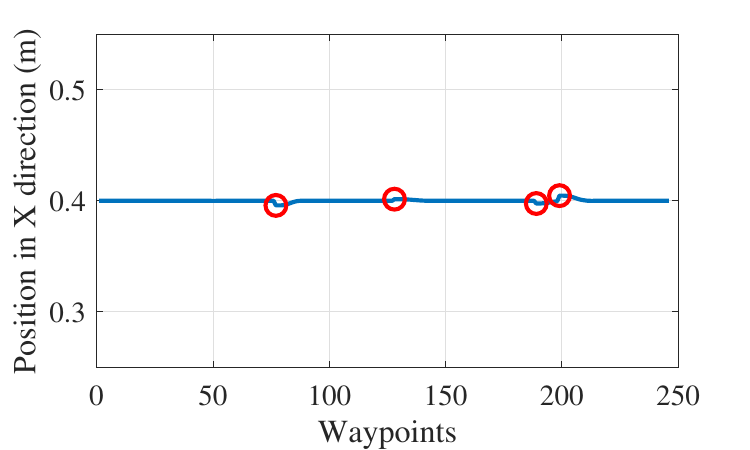}
	\label{fig:errors_x}
	}
	\subfigure[The position of the end-effector in the Y direction]{
	\includegraphics[width=0.62\columnwidth]{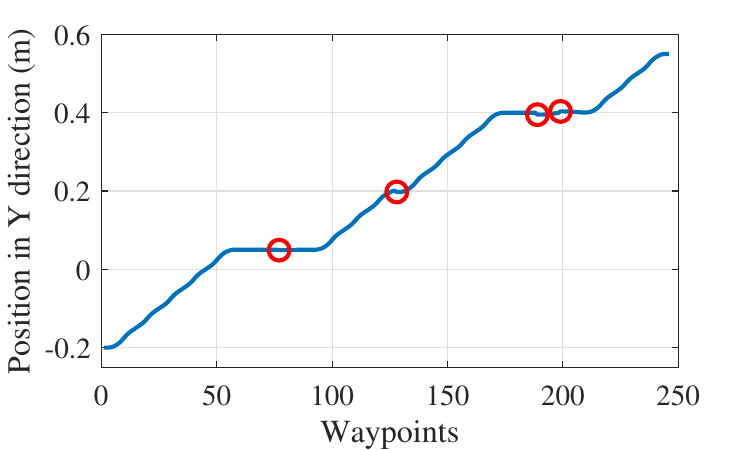}
	\label{fig:errors_y}
	}
    \subfigure[The position of the end-effector in the Z direction]{
	\includegraphics[width=0.62\columnwidth]{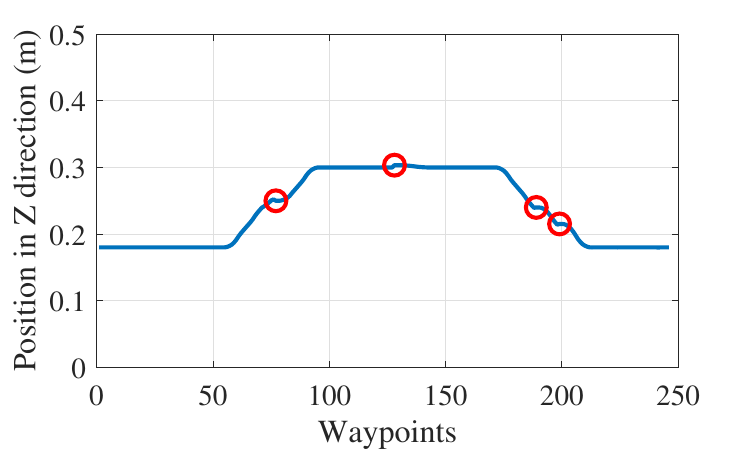}
	\label{fig:errors_z}
    }
	\caption{Precision analysis after re-configuration in the experimental test scenario B: The end-effector's position after re-configuration is highlighted with red circles.}
	\vspace{5pt}
	\label{fig:positions}
\end{figure*}

\begin{table*}[!]
\begin{center}
	\begin{tabular}{clclclc}
\toprule[1.5pt]
Waypoints  &  ~~~~Target positions (x,y,z)   & Actually positions (x,y,z) & ~~Error \\
\toprule[1.5pt]
     1& (0.4000m, 0.0500m, 0.2500m)& (0.3958m, 0.0489m, 0.2499m)& 0.0043m\\
        2& (0.4000m, 0.2000m, 0.3000m)& (0.4014m, 0.1975m, 0.3035m)& 0.0045m\\
        3& (0.4000m, 0.4000m, 0.2400m)& (0.3975m, 0.3949m, 0.2401m)& 0.0057m\\

		 4& (0.4000m, 0.4000m, 0.2100m)& (0.4046m, 0.4032m, 0.2152m)& 0.0076m\\
		\toprule[1.5pt]
	\end{tabular}
	\vspace{5pt}
	\caption{Details of waypoints' positions after re-configurations}
	\label{tab:waypoints_detail}
 \end{center} \vspace{-10pt}
\end{table*}

Furthermore, the process of OMRM includes finding configuration candidates from the database, ranking candidates based on RMSE, and filtering candidates by collision-checking. The average solution time of OMRM is presented in Table~\ref{tab:OMRM_detail}. In general, considering that the database builds the relationship between the task space and the configuration space offline, our OMRM only requires a cheap selection cost. 

\section{Experimental tests}
This section presents experimental tests to validate the effectiveness of the proposed motion planning algorithm. An experiment platform has been built up and several experimental tests have been conducted.

\subsection{Experiment setup}
A UR5e robot arm attaching a Robotiq 2F-85 gripper is employed to conduct the experimental tests. To plan the end-effector path, a 3D grid map, in which the length, width, and height are all $1m$ is constructed along with $X$, $Y$, and $Z$ directions. Each cell of the grid map employed by the A* algorithm is defined as a cubic volume with a side length of $0.02m$. This grid resolution has been chosen to be suitable for facilitating effective collision avoidance during the planning phase of the manipulator's end-effector.
We employ a Vicon motion capture system for the purpose of perception and tracking the positions of moving obstacles. The system is comprised of multiple high-speed cameras positioned to capture the specific region necessitating motion tracking. Moving obstacles are equipped with gray markers, that are designed to reflect the infrared light emitted by these cameras. 
By comparing the positions of the markers in multiple camera views, it can precisely reconstruct the obstacle's movement in three dimensions (3D). Additionally, the obstacle models incorporated into the collision checking process are intentionally designed with enlarged dimension relative to their original size. This scaling is implemented to ensure a safety distance between the obstacles and the manipulator, even in scenarios where a collision is detected. Note that we implement safety mechanisms as a precautionary measure in our experimental tests to protect the manipulator, once self-collisions or singularities are detected.

Additionally, we didn't explicitly specify the orientation of the manipulator during the planning phase. There are two reasons for not considering the orientation in our studies. First, the specific application scenario under investigation in this paper is disassembly processes with human, where the manipulator moves to desired orientations to manipulate disassembled components. It remains feasible to adjust the manipulator's orientation after it safely reaches the desired position. Secondly, specifying orientation of the manipulator during the planning phase will bring more computational efforts, especially considering that the planner aims to find a path that minimizes the end-effector's path with minimum reconfiguration efforts. Introducing orientation constraints may increase the planning time, potentially impeding the manipulator's ability to attain a collision-free configuration while reaching to the desired end-effector's position.

\subsection{Experiment scenario A}
Fig.~\ref{fig:expA} demonstrates the experimental test scenario A. The start position is denoted as $x_{0}{=}(0.44m,0.08m,0.42m)$, while the goal position is $x_{K+1}{=}(0.44m, 0.44m, 0.42m)$. The planned path for the end-effector is a straight red line below the human arm. In this case, the gripper, along with some links of the UR5e robot arm, has to move below the human arm. Consequently, relying solely on the planning in the task space, as represented by the red line, is insufficient. This necessitates the activation of our OMRM to provide an alternative configuration and enable the robot to successfully reach the target position after the re-configuration. 
Fig.~\ref{fig:expA} (b) and (c) show the experimental frames right before and after the manipulator reconfiguration, respectively.

\subsection{Experiment scenario B}
Fig.~\ref{fig:expB} demonstrates the experimental test scenario B which requires multiple re-configurations to reach the target position. The start position is $x_{0}{=}(0.40m, -0.20m,0.18m)$, while the goal position is $x_{K+1}{=}(0.40m, 0.56m, 0.18m)$. The A* algorithm is employed to re-plan the path of the end-effector, adapting to the dynamic position of the box. The average planning time for the planning is around $33.98ms$. The manipulator follows the re-planned path in the task space. Fig.~\ref{fig:expB} (c) and (f) show that the human arm moves toward the manipulator, and our OMRM is triggered based on the sudden interventions of the human arm, which is shown in Fig.~\ref{fig:expB} (d) and (g). Fig.~\ref{fig:expB} (e) and (h) illustrates the moments right after the first and second re-configurations. Furthermore, the manipulator needs the third and fourth re-configurations to avoid the box, and no collision is detected after Fig.~\ref{fig:expB} (l).

The recorded positions of the end-effector during the experimental test scenario B are illustrated in Fig.~\ref{fig:positions}. Besides the inherent error stemming from the mechanical design, control system, and actuators of the manipulator, the actual end-effector's path also incorporates a tolerance-induced error resulting from manipulator re-configurations. 
Detailed positional information of four target points after re-configurations are presented in Table~\ref{tab:waypoints_detail}.
It is worth noting that there exists a capacity for manually reducing the positional error by specifying a smaller error tolerance $\zeta$. 
A smaller error tolerance correlates with a finer level of precision for the end-effector's position after a re-configuration. 
However, this improvement of precision necessitates a trade-off, as the reduction in error tolerance corresponds to a reduction in the count of available configuration candidates.
 In this study, we choose the tolerance $\zeta$ as 0.01m to ensure a sufficient number of configuration candidates for each waypoint cell. Importantly, this selected tolerance of 0.01m remains reasonable when compared to the overall length of the planned end-effector's path, i.e., a total of 1.00m. Furthermore, the errors shown in Table~\ref{tab:waypoints_detail} consistently remain below the prescribed threshold during both simulation and experimental tests.

\section{Conclusions and Future Work}

This study aims to efficiently plan the motion for manipulators when executing tasks as well as satisfying certain requirements, e.g., following the planned path and avoiding collision with human operators. To this end, we first formulate the motion planning problem into an optimization problem with constraints, and then solve the optimization problem using $A^*$ algorithm and OMRM. To enable the end-effector to reach the goal position, a path planning algorithm based on $A^*$ plans the collision-free end-effector path with the minimum distance. To follow the reference path and avoid the collision of the manipulator body, OMRM relying on a database selects an alternative configuration to replace the current one when facing a collision. The database maps the task space to the configuration space offline, which reduces the computational time. The proposed algorithm has been validated in two different scenarios, and the results show that the manipulator follows the planned path and efficiently avoids the human arm successfully in both scenarios.

Considering that the construction of the configuration database ignores the self-collisions and singularities of the manipulator, future studies can prioritize developing algorithms to eliminate these dangerous configurations from the database. Meanwhile, additional safety mechanisms during experimental tests can be implemented to enlarge the safety margin. Furthermore, given that minimizing the joint-angle change may result in frequent re-planning even when the human barely moves, we intend to address this potential issue through two proposed approaches. The first approach is to explore a trade-off problem between the joint-angle change and the minimum safety distance. Rather than employing a fixed safety distance between the human and the manipulator, an adaptive safety distance could be dynamically defined based on the effort required for joint-angle changes. This adaptive distance could be utilized in the planning process to mitigate the need for unnecessary re-plannings. The second approach is to leverage predictive models of human motions and incorporating them into the safe manipulator motion planning could offer a solution. By anticipating the human motions, the manipulator could proactively adjust its configuration to avoid the frequent re-planning caused by close proximity to the human.

\bibliographystyle{IEEEtran}
\bibliography{Reference}{}

\end{document}